\documentclass{article} % For LaTeX2e
\usepackage{iclr2027_conference,times}

\usepackage{amsmath,amsfonts,bm}

\def\eqref#1{equation~\ref{#1}}
\def\1{\bm{1}}

\def\vmu{{\bm{\mu}}}

\def\vf{{\bm{f}}}

\def\vk{{\bm{k}}}

\def\vm{{\bm{m}}}

\def\vq{{\bm{q}}}
\def\vr{{\bm{r}}}

\def\vv{{\bm{v}}}

\def\vx{{\bm{x}}}

\DeclareMathAlphabet{\mathsfit}{\encodingdefault}{\sfdefault}{m}{sl}
\SetMathAlphabet{\mathsfit}{bold}{\encodingdefault}{\sfdefault}{bx}{n}

\usepackage{hyperref}
\usepackage{url}
\usepackage{booktabs}
\usepackage{array}
\usepackage{graphicx}
\usepackage{amsmath}
\usepackage{amssymb}
\usepackage{xcolor}
\usepackage{colortbl}
\usepackage{multirow}
\usepackage{algorithm}
\usepackage{algpseudocode}
\usepackage{subcaption}
\usepackage{tikz}
\usetikzlibrary{arrows.meta,positioning,calc,decorations.pathreplacing}
\definecolor{OursFill}{HTML}{FDE3D2}
\definecolor{OursEdge}{HTML}{D55E00}
\definecolor{Spill}{HTML}{E8A0A0}

\graphicspath{{figures/}}
\fancypagestyle{firstpagestyle}{%
  \fancyhf{}  
  \fancyhead[L]{\includegraphics[height=0.66cm]{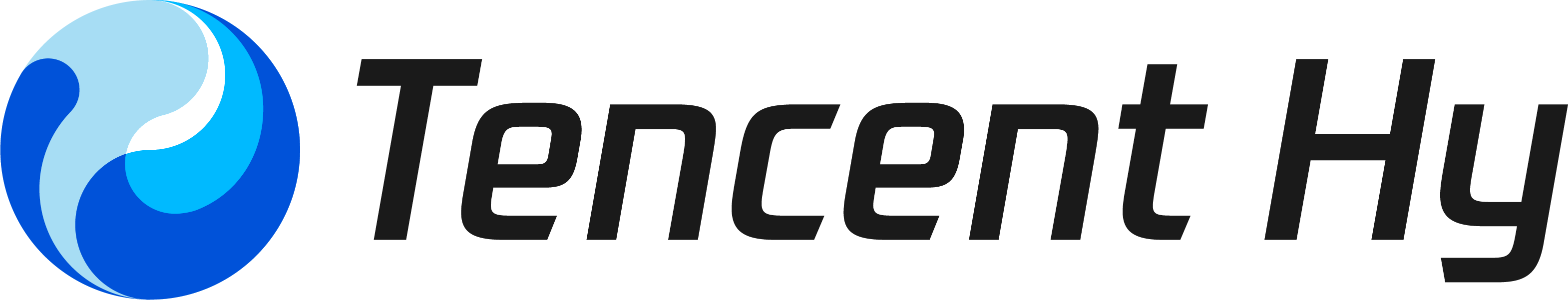}}
  \renewcommand{\headrulewidth}{0.8pt}  
  \renewcommand{\footrulewidth}{0pt}
}

\title{D-Quant: Driftable Entropy Coding for KV Cache Quantization}

\iclrfinalcopy % Uncomment for camera-ready version, but NOT for submission.

\makeatletter
\long\gdef\affiliations#1{\gdef\@affiliations{#1}}
\gdef\@affiliations{}
\def\@maketitle{\vbox{\hsize\textwidth \centering
  {\fontsize{14}{16}\selectfont\bfseries \@title \par}
  \vskip 0.14in
  {\fontsize{12}{14}\selectfont\bfseries \@author \par}
  \vskip 0.09in
  {\fontsize{10}{12}\selectfont \@affiliations \par}
  \vskip 0.18in}}
\makeatother

\author{
    Yi Su \quad
    Hong Liu \quad
    Guanghua Yu\thanks{Corresponding author.} \quad
    Jianchen Zhu
}
\affiliations{
    Tencent \\
    \texttt{brunosu@tencent.com} \quad \texttt{lucayu@tencent.com}
}

\begin{document}

\maketitle
\thispagestyle{firstpagestyle} 
\begin{abstract}
The KV cache has become a major bottleneck in deploying LLMs, as its memory footprint grows linearly with sequence length and batch size, imposing substantial pressure on both memory capacity and bandwidth.
Among various KV cache compression techniques, quantization is particularly attractive due to its effectiveness and ease of deployment.
However, most existing methods rely on fixed-width quantization, where a $b$ bit representation is inherently limited to $2^b$ quantization levels.
As the bit width decreases, the number of available levels shrinks exponentially, leading to severe information loss and rapid performance degradation.
We further observe that fixed-width quantization fails to exploit the highly non-uniform distribution of KV cache.
After rotation and normalization, KV values approximately follow a normal distribution, with most values concentrated near the center and only a small fraction appearing in the tails.
Nevertheless, fixed-width coding allocates the same number of bits to frequent and rare symbols.
Entropy coding naturally exploits such non-uniformity by assigning shorter codewords to frequent symbols and longer ones to rare symbols, substantially reducing the average number of bits required for representation.
However, its variable-length output is not suited to highly parallel attention kernels, where efficient dequantization and computation rely on regular memory layouts and fixed-stride accesses.
To bridge this gap, we propose \textbf{D-Quant}, a flexible KV cache quantization framework that introduces a \textbf{drift} mechanism to convert entropy-coded representations of each token into fixed-size bitstreams, enabling regular memory access and parallel dequantization within attention kernels.
D-Quant also provides greater flexibility than conventional quantization methods, allowing the bit budget and number of quantization levels to be configured more freely rather than being constrained by fixed bit widths and group sizes.
Furthermore, since the rotated and normalized KV values closely follow a normal distribution, the probability model for entropy coding can be directly derived from the standard normal distribution, making D-Quant entirely calibration-free.
Extensive experiments show that D-Quant preserves near BF16 performance at only $2.26$ bits per element while substantially outperforming existing baselines.
D-Quant further achieves up to a $7\times$ reduction in KV cache memory footprint and a $3.5\times$ improvement in decoding throughput.
\end{abstract}
\begin{figure}[h]
\centering
\resizebox{0.86\textwidth}{!}{% Encoding one token, and what drift does to make it fit.
% Included by paper/4-Method.tex. Needs arrows.meta, positioning, calc and
% decorations.pathreplacing, all loaded in main.tex.
\begin{tikzpicture}[
  font=\footnotesize,
  box/.style={draw=black!50, rounded corners=1.6pt, minimum height=8.4mm,
              minimum width=13mm, inner xsep=3pt, align=center, fill=white,
              line width=0.45pt},
  ours/.style={box, draw=OursEdge, fill=OursFill, line width=1.0pt},
  off/.style={box, draw=black!25, dashed, text=black!45, line width=0.45pt},
  ar/.style={-{Latex[length=1.6mm, width=1.2mm]}, black!50, line width=0.45pt},
  lbl/.style={font=\footnotesize, text=black!60},
  side/.style={font=\small\bfseries, text=black!70},
]

% ---------------- pipeline ----------------
\node[box, minimum width=7mm] (kin) at (0,0) {$\vk_t$};
\node[box,  right=3mm of kin]  (krot) {Hadamard};
\node[box,  right=3mm of krot] (kmu)  {remove\\[-2pt]channel mean};
\node[box,  right=3mm of kmu]  (kgr)  {affine\\[-2pt]quantization};
\node[ours, right=3mm of kgr]  (kdr)  {drift};
\node[ours, right=3mm of kdr]  (kra)  {rANS};
\node[box,  right=3mm of kra]  (kout) {container};
\foreach \a/\b in {kin/krot, krot/kmu, kmu/kgr, kgr/kdr, kdr/kra, kra/kout}
  \draw[ar] (\a) -- (\b);

\node[box, minimum width=7mm, below=7mm of kin] (vin) {$\vv_t$};
\node[off] (vrot) at (krot |- vin) {Hadamard\\[-2pt](offline)};
\node[box] (vgr)  at (kgr  |- vin) {affine\\[-2pt]quantization};
\node[ours](vdr)  at (kdr  |- vin) {drift};
\node[ours](vra)  at (kra  |- vin) {rANS};
\node[box] (vout) at (kout |- vin) {container};
\draw[ar] (vin) -- (vrot);
\draw[ar] (vrot) -- (vgr);
\foreach \a/\b in {vgr/vdr, vdr/vra, vra/vout} \draw[ar] (\a) -- (\b);

\draw[decorate, decoration={brace, amplitude=3pt}, black!40]
  ($(kin.north west)+(0,1.6mm)$) -- ($(kgr.north east)+(0,1.6mm)$)
  node[midway, above=1.6pt, lbl] {preprocessing};
\draw[decorate, decoration={brace, amplitude=3pt}, OursEdge!70]
  ($(kdr.north west)+(0,1.6mm)$) -- ($(kout.north east)+(0,1.6mm)$)
  node[midway, above=1.6pt, lbl, text=OursEdge!85!black] {entropy coding};

\end{tikzpicture}}
\caption{\textbf{Overview of our method.} A stream is the Keys or Values of one token and shares a single scale and offset.
After preprocessing (rotation, mean removal, affine grid) and affine quantization, the symbol indices are entropy-coded into a container of fixed byte length, and a length-constrained assignment makes every stream fit that container exactly.}
\label{fig:pipeline}
\end{figure}
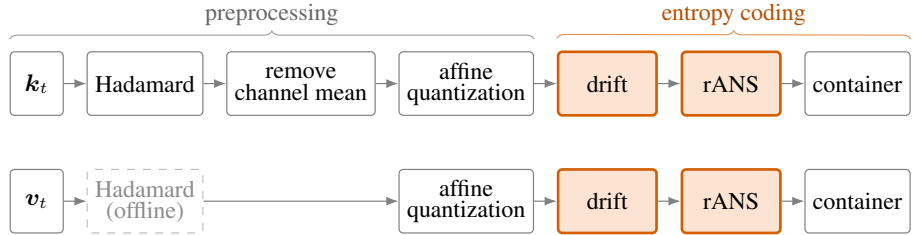
\section{Introduction}
\label{sec:intro}

As Large Language Models (LLMs) become increasingly capable, they are being widely adopted in applications such as multi-turn conversations, complex reasoning, and coding agents \citep{intro1,intro2,intro3}. These applications place growing demands on long-context capabilities, leading to substantial resource overhead during deployment.
During auto-regressive inference, the KV cache of every processed token must be retained for subsequent generation. Unlike model weights, whose size remains fixed, the KV cache grows linearly with sequence length and batch size, imposing substantial pressure on both memory capacity and bandwidth.
Quantization, which has proven effective and deployment-friendly for compressing model weights and activations \citep{frantar2023gptq,xiao2023smoothquant,lin2024awq}, therefore emerges as a promising solution to this growing KV cache bottleneck.

Existing KV cache quantization methods primarily improve quantization granularity or preprocessing strategies, but typically encode quantized values using a fixed number of bits \citep{liu2024kivi,hooper2024kvquant,ashkboos2024quarot,su2026oscar}.
Under this paradigm, each value is represented using $b$ bits and can therefore take only $2^b$ quantization levels.
As the bit width decreases, the number of available levels drops exponentially, substantially limiting representation capacity and ultimately degrading quantization accuracy.
Fixed-width quantization also fails to exploit the non-uniform distribution of KV cache. Existing methods such as rotation \citep{ashkboos2024quarot,tseng2024quip} and normalization \citep{su2026oscar} have developed effective preprocessing techniques to make KV caches more amenable to quantization. 
We find that KV cache closely follow a normal distribution after preprossing, with most values concentrated near the center and only a small fraction appearing in the tails.
However, fixed-width quantization cannot exploit this distributional structure because it assigns the same number of bits to frequent and rare symbols.
Entropy coding provides a natural way to exploit this non-uniformity.
By assigning shorter codewords to frequent symbols and longer ones to rare symbols, it reduces the average number of bits required to represent a given sequence.
However, entropy coding produces variable-length bitstreams, making it difficult to integrate with highly parallel attention kernels.
Efficient dequantization and computation in these kernels typically rely on regular data layouts and fixed-stride accesses.

To address this issue, we propose \textbf{D-Quant}, a flexible KV cache quantization framework that introduces a \textbf{drift} mechanism to convert entropy-coded representations of each token into fixed-size bitstreams, enabling regular memory access and parallel dequantization within attention kernels.
To satisfy this constraint with minimal accuracy loss, we formulate quantization as a length-constrained assignment problem that minimizes reconstruction error under the target bit budget. When the encoded stream exceeds the budget, a small fraction of values drift from high-cost symbols to nearby lower-cost ones until the length constraint is met. In this way, D-Quant adapts only the necessary values rather than uniformly reducing the precision of the entire representation. We solve this minimization problem efficiently using a Lagrangian formulation with bisection.
This design also makes D-Quant substantially more flexible than conventional fixed-width quantization. The storage budget is no longer tied to an integer bit width or a particular group size, and the number of quantization levels is no longer restricted to $2^b$. For example, at a matched storage budget of $2.26$ bits, D-Quant uses eight levels for keys and six for values. The fixed-container formulation further allows the storage budget to be flexibly divided between keys and values according to their different sensitivity to quantization.
Moreover, entropy coding typically requires a probability model estimated from the online data. However, rotation and normalization have made the KV distribution closely approximate a standard normal distribution. 
Therefore, we directly derive the probability model from the standard normal distribution, eliminating the need for calibration while retaining the benefits of distribution-aware entropy coding.

Our contributions are summarized as follows:
\begin{itemize}
    \item We identify a fundamental limitation of fixed-width KV cache quantization. It couples the number of quantization levels to the bit width and fails to exploit the non-uniform distribution of KV cache.
    \item We propose D-Quant, which combines entropy coding with fixed-size per-token storage through a drift mechanism, enabling flexible bit budgets and quantization levels while preserving efficient parallel processing. D-Quant further derives its probability model directly from the standard normal distribution, requiring no calibration data.
    \item Extensive experiments show that D-Quant maintains near BF16 performance at 2.26 bits per value, substantially outperforming existing baselines while reducing KV cache memory by up to $7\times$ and improving decoding throughput by up to $3.5\times$.
\end{itemize}
\section{Related Work}
\label{sec:related}

\paragraph{KV cache quantization.}
KV cache quantization has been widely explored to reduce the memory and bandwidth overhead of long-context inference.
KIVI \citep{liu2024kivi} identifies distinct distributional properties of keys and values, motivating per-channel quantization for keys and per-token quantization for values.
KVQuant \citep{hooper2024kvquant} improves low-bit quantization through non-uniform quantization and explicit outlier handling, while OTT \citep{su2025accurate} preserves salient outlier tokens in higher precision to mitigate quantization error.
Atom \citep{zhao2024atom} reorganizes channels to better isolate outliers, while QuaRot \citep{ashkboos2024quarot} applies Hadamard rotations to suppress outliers and improve quantizability.
More recent work continues to push KV cache quantization toward lower bit widths through improved preprocessing, quantization schemes, and dedicated system optimizations \citep{su2026oscar}.
Complementary to these efforts, our work explores the largely overlooked coding format of quantized KV values.

\paragraph{Entropy coding.}
For non-uniform data, entropy coding reduces the expected code length by assigning shorter codewords to more frequent symbols and longer ones to less frequent symbols, with the source entropy providing a fundamental lower bound on the average number of bits required per symbol \citep{shannon1948}.
Huffman coding \citep{huffman1952} constructs prefix codes whose average code length is within one bit of the entropy bound. Arithmetic coding \citep{witten1987arithmetic} approaches this bound more closely by encoding an entire sequence into a single interval. Asymmetric numeral systems (ANS) \citep{duda2013ans} achieve compression efficiency comparable to arithmetic coding while supporting efficient table-based implementations, making them well suited for high-throughput decoding.
When entropy coding is combined with lossy quantization, the coding cost of each symbol becomes part of the quantization objective, commonly formulated as a optimization that balances reconstruction error against coding rate.
Such formulations have been extensively studied in image and video compression \citep{ortega1998rdmethods,sullivan1998rdo,marcellin1990tcq,marpe2003cabac} and neural network compression \citep{wiedemann2020deepcabac}.
A standard approach uses a Lagrangian objective to jointly account for reconstruction error and code length when assigning quantization symbols.
Entropy coding has also been explored for KV cache compression.
CacheGen \citep{liu2024cachegen} encodes KV caches into compact bitstreams to reduce communication overhead during KV cache streaming.
SplitZip \citep{guo2026splitzip} employs lightweight lossless compression for KV cache transfer in disaggregated inference.
These methods primarily target KV cache communication, where compressed streams can be fully decoded before attention computation.
In contrast, directly operating on entropy-coded KV caches within attention kernels is more challenging, as variable-length representations introduce irregular memory access and hinder efficient parallel decoding \citep{jiang2025kvcomp}.
\section{Preliminary}
\label{sec:prelim}

\subsection{Background}

\paragraph{Group Quantization.}
Let $\vx \in \mathbb{R}^{n}$ denote the key or value vector of a single token at one layer. Group-wise quantization partitions $\vx$ into groups of $g$ values, each sharing a scale $s$ and zero point $o$. Under $b$-bit affine quantization, each value is reconstructed as
\begin{equation}
    \hat{x}_i = o + s m_i, \qquad m_i \in \{0,\ldots,2^b-1\},
    \label{eq:affine}
\end{equation}
where $m_i$ is the quantized \emph{symbol}. A fixed-width $b$-bit representation therefore provides exactly $2^b$ reconstruction levels. We report storage in bits per value (BPV), including quantized symbols and metadata. For example, INT2 with an FP16 scale and zero point per $128$ values requires 2.25 BPV.

\paragraph{Entropy Coding.}
Entropy coding assigns shorter representations to more probable symbols. For a symbol $m$ with probability $p_m$, the ideal coding cost is $r(m)=-\log_2 p_m$, whose expectation is the entropy $H(p)$. Thus, an alphabet of $M$ symbols can require substantially fewer than $\log_2 M$ bits per value when its distribution is non-uniform, allowing more quantization levels under the same average storage budget.

We use range asymmetric numeral systems (rANS) \citep{duda2013ans} for efficient entropy coding. Given precision $p$, each symbol $m$ is assigned an integer frequency $f_m$ and cumulative frequency $c_m$, with $\sum_m f_m=2^p$ and $p_m=f_m/2^p$. Encoding and decoding update the integer state $x$ as
\begin{equation}
\begin{aligned}
\mathrm{Enc}(x,m) &=
    \left\lfloor \frac{x}{f_m}\right\rfloor 2^p
    +(x\bmod f_m)+c_m,\\
\mathrm{Dec}(x,m) &=
    f_m\left\lfloor\frac{x}{2^p}\right\rfloor
    +(x\bmod 2^p)-c_m.
\end{aligned}
\label{eq:rans}
\end{equation}
During decoding, $z=x\bmod2^p$ uniquely identifies the symbol whose interval $[c_m,c_m+f_m)$ contains $z$. This mapping can be implemented by a $2^p$-entry lookup table; with $p=8$, only $256$ entries are required. We store the corresponding reconstruction values in the same table, allowing decoding and dequantization to share a lookup. Since rANS decodes symbols in reverse order, each stream is encoded backwards so that decoding proceeds in its original order.

Unlike fixed-width coding, rANS allows coding costs to vary across symbols, decoupling the number of quantization levels from a fixed per-value bit width. The resulting stream length, however, varies with its symbols. D-Quant exploits the former while eliminating the latter by enforcing a fixed storage budget for every stream.

\subsection{Observations}
\label{sec:observations}

\paragraph{Preprocessed KV caches exhibit predictable and compressible distributions.}

Entropy coding is effective when symbols follow non-uniform distributions.
We investigate these properties on conventionally
processed KV caches.
Specifically, we quantize each token into an eight-level
per-token grid and measure three quantities: reconstruction error (NMSE), the
potential entropy coding gain over fixed-width coding
($\mathrm{Coding\ Gain}=\log_2 M-H$), and the excess coding rate of using a
standard-normal-derived model instead of an empirical model
($\mathrm{Normal\ Gap}=\mathrm{CE}-H$).

\begin{table}[t]
\caption{Effect of preprocessing on quantization error and entropy modeling.
Results are measured on captured Qwen3-8B. Each token
is quantized into an eight-level per-token grid. Normal Gap measures the excess
rate of using a standard-normal-derived entropy model instead of an empirical
model, while Coding Gain measures the potential saving over fixed-width quantization.}
\label{tab:stages}
\centering
\setlength{\tabcolsep}{4pt}
\resizebox{\textwidth}{!}{
\begin{tabular}{lcccccccc}
\toprule
& \multicolumn{4}{c}{Key} & \multicolumn{4}{c}{Value} \\
\cmidrule(lr){2-5}\cmidrule(lr){6-9}
Stage 
& NMSE 
& Mean/RMS 
& Normal Gap 
& Coding Gain
& NMSE 
& Mean/RMS 
& Normal Gap 
& Coding Gain \\
\midrule
Raw              
& $180.9\%$ 
& $0.367$ 
& $0.284$ 
& $1.47$
& $14.0\%$ 
& $0.250$ 
& $0.052$ 
& $1.01$
\\

$+$ rotation     
& $2.45\%$  
& $0.662$ 
& $0.078$ 
& $0.48$
& $9.40\%$ 
& $0.302$ 
& $0.008$ 
& $0.85$
\\

$+$ mean removal 
& $0.48\%$  
& $0.009$ 
& $0.0005$ 
& $0.72$
& $7.89\%$ 
& $0.003$ 
& $0.009$ 
& $0.86$
\\
\bottomrule
\end{tabular}}
\end{table}

\begin{figure}[t]
\centering
\includegraphics[width=\textwidth]{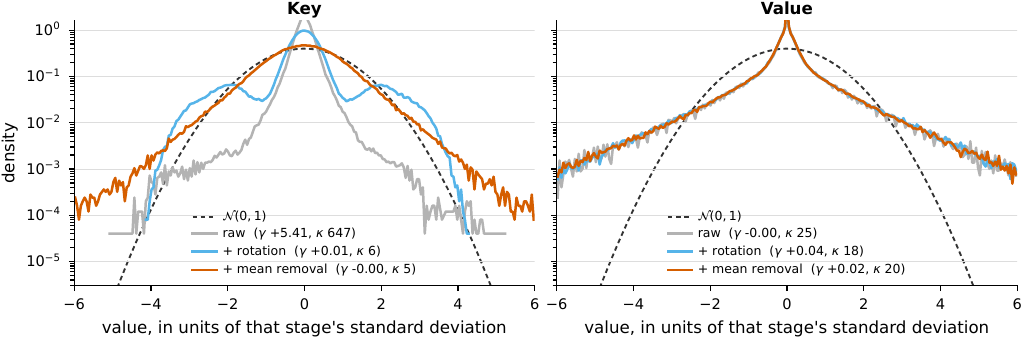}
\caption{Distribution of KV values before and after preprocessing. Raw KV
tensors exhibit strong outliers and non-Gaussian patterns. Hadamard rotation
substantially reduces distribution irregularity, and mean removal further
centers key distributions.}
\label{fig:observation_shapes}
\end{figure}

Hadamard rotation spreads structured outliers across channels and makes KV
distributions substantially closer to Gaussian
(Figure~\ref{fig:observation_shapes}). However, rotation alone does not remove
the channel-wise mean of keys, which cannot be captured by a token-level scalar
offset. Removing this residual mean further reduces key NMSE to $0.48\%$ and,
more importantly, reduces the Normal Gap to only $5\times10^{-4}$ bits/value.
This indicates that the probability model required by entropy coding can be
accurately derived from a standard normal distribution without calibration
data. Meanwhile, the processed KV cache still retains substantial entropy
coding potential, achieving $0.72$ and $0.86$ bits/value Coding Gain for keys and
values, respectively.

\paragraph{Entropy coding decouples quantization levels from storage budgets.}

Fixed-width quantization inherently couples the representation capacity with
the storage budget. A $b$-bit representation provides exactly $2^b$
quantization levels. Reducing the bit width therefore inevitably reduces the
available alphabet. Entropy coding changes this constraint by allocating bits
according to symbol probabilities, where the storage budget limits the total
coding rate rather than the number of quantization levels.
To study this effect, we fix the storage budget of each token and vary the
number of quantization levels. For each alphabet size, we solve the
corresponding rate-distortion optimization under the same storage constraint and
measure the resulting reconstruction error.

\begin{table}[t]
\caption{Reconstruction error under different alphabet sizes with fixed storage
budgets. The key and value containers are fixed to 2.68 BPE and 1.84 BPE,
respectively. Dashes indicate that the corresponding alphabet cannot satisfy
the storage constraint.}
\label{tab:levels}
\centering
\small
\begin{tabular}{lcccccccc}
\toprule
Levels $M$
& 4 & 6 & 7 & 8 & 10 & 11 & 12 & 14 \\
\midrule
Key, 2.68 bit
&18.62\%
&9.81\%
&7.47\%
&5.93\%
&4.51\%
&4.32\%
&4.33\%
&4.98\%
\\

Value, 1.84 bit
&21.53\%
&14.12\%
&13.25\%
&14.11\%
&29.64\%
&80.48\%
&--
&--
\\
\bottomrule
\end{tabular}
\end{table}

Table~\ref{tab:levels} shows that entropy coding provides substantially more
flexibility under a fixed storage budget. The number of quantization levels is
no longer constrained by the bit width, but can be selected according to the
distribution of KV values and the desired rate-distortion trade-off. This
decoupling enables more expressive representations while preserving a fixed
memory layout.
\section{Method}
\label{sec:method}

\subsection{Overview}

D-Quant compresses each key or value tensor of one token into a fixed-size container. As illustrated in Figure~\ref{fig:pipeline}, the encoding pipeline consists of two stages.
First, we apply lightweight preprocessing to improve the distribution of KV values, including Hadamard rotation, mean removal for keys, and affine quantization.
Second, instead of representing quantized symbols with fixed-width codes, D-Quant applies entropy coding to exploit the non-uniform symbol distribution.

A direct application of entropy coding, however, produces variable-length representations, which prevents fixed-stride memory access and efficient parallel decoding in attention kernels.
D-Quant addresses this issue by introducing a fixed-size container for each stream and a drift mechanism that optimizes symbol assignments under the container constraint.
The resulting representation preserves the compression benefits of entropy coding while maintaining the regular memory layout required by efficient KV cache dequantization.

\subsection{Fixed-size entropy coding for KV cache}
\label{sec:container}

Instead of assigning a fixed-width codeword to each value, D-Quant allocates a container of $C$ bytes per stream and encodes its quantized symbols using entropy coding. A stream denotes the $n$ values of one key or value tensor corresponding to a single token. The encoded stream satisfies
\begin{equation}
    \sum_{i=1}^{n} r(m_i) \le B,
    \qquad B=8C-\sigma ,
    \label{eq:budget}
\end{equation}
where $r(m_i)$ is the entropy-coded length of symbol $m_i$ and $\sigma$ denotes the final state overhead of the entropy coder. Since each stream occupies a predetermined storage region, the KV cache remains a regular array of fixed-stride records, enabling efficient parallelism inside attention kernels.

Before entropy coding, D-Quant applies lightweight preprocessing to obtain a predictable symbol distribution. Following prior KV cache quantization methods \citep{ashkboos2024quarot,liu2024kivi}, we apply Hadamard rotation to each attention head. For values, the rotation can be folded into projection weights offline, while for keys it is applied online due to the non-commutativity with RoPE \citep{su2024rope}. Keys further remove a windowed channel-wise mean. The processed stream is quantized with a token-wise affine:
\begin{equation}
    \hat{x}_i=o+s m_i,\qquad m_i\in\{0,\dots,M-1\},
    \label{eq:affine_token}
\end{equation}
where $s=(\max(\vx)-\min(\vx))/(M-1)$ and $o=\min(\vx)$.
The entropy model is analytically derived from Gaussian order statistics and requires no calibration data. Due to the affine invariance of normalized positions, the same probability table can be shared across different tokens and layers.

\subsection{Drift: length-constrained assignment}
\label{sec:drift}

Although entropy coding provides flexible representations, the resulting code length depends on the assigned symbols and may exceed the predefined budget. Instead of reducing the number of quantization levels, D-Quant directly optimizes the symbol assignment under the length constraint:
\begin{equation}
    \min_{\vm\in\{0,\dots,M-1\}^{n}}
    \sum_{i=1}^{n}
    \left(x_i-(o+s m_i)\right)^2
    \quad
    \mathrm{s.t.}\quad
    \sum_{i=1}^{n} r(m_i)\le B ,
    \label{eq:drift}
\end{equation}
where $r(m_i)$ is the entropy-coded length obtained from the probability model.
We solve this constrained optimization through a Lagrangian formulation:
\begin{equation}
    \min_{\vm}
    \sum_{i=1}^{n}
    \left[
    D_i(m_i)+\lambda r(m_i)
    \right],
    \label{eq:lagrangian}
\end{equation}
where $D_i(m_i)$ denotes the reconstruction error of assigning value $x_i$ to level $m_i$. For a fixed multiplier $\lambda$, the objective is separable across values, allowing each symbol assignment to be independently optimized. We use bisection to find the multiplier that satisfies the code-length constraint.

Intuitively, D-Quant does not uniformly reduce the representation precision when the encoded stream exceeds its budget. Instead, only a small fraction of values assigned to expensive symbols are allowed to \emph{drift} toward nearby lower-cost symbols. This progressively reduces the total code length with minimal reconstruction error, preserving the benefits of a large entropy-coded alphabet. The resulting assignment jointly optimizes reconstruction quality and coding cost, enabling every stream to satisfy its fixed container constraint.

\subsection{Practical refinements}
\label{sec:refinements}
\paragraph{Separate K/V configurations.} Keys and values play different roles in attention and exhibit different sensitivities to quantization. Errors in keys directly perturb the attention logits and can be amplified through the exponential operation in softmax, whereas errors in values are introduced after softmax and are aggregated by the attention weights. Empirically, we find keys to be more sensitive to quantization and therefore allocate them a larger container budget. D-Quant further allows the number of quantization levels for keys and values to be configured independently.

\paragraph{Distribution sharpening.} The analytic probability model describes the symbol distribution before drift, whereas drift preferentially moves values toward lower-cost, more probable symbols, making the final coded distribution more concentrated. We account for this shift by sharpening the analytic probabilities as $p_m' \propto p_m^{\alpha}$ with $\alpha>1$ before constructing the entropy table. A single $\alpha$ is used across all models and layers.

\subsection{Kernels}
\label{sec:kernel}

\paragraph{Encoding.} A straightforward implementation of the drift solver materializes an $n\times M$ distortion matrix at every bisection step, introducing substantial memory traffic for values that are inexpensive to recompute. We instead assign one warp to each stream and keep its values on chip throughout the optimization. At each bisection step, distortion is recomputed on the fly for the $M$ candidate levels, while the resulting code length is reduced through warp-level operations. The final affine refit and rANS encoding are performed within the same warp, avoiding intermediate materialization and minimizing memory traffic.

\paragraph{Decoding.} Since each entropy-coded stream spans all KV heads of a token, conventional head-centric attention kernels would repeatedly traverse the same stream when different heads consume their corresponding values. We therefore organize decoding around tokens: each thread block owns a block of tokens, decodes each stream only once, and directly feeds the decoded levels into the corresponding attention computation. As the stream is decoded one KV head at a time, warps partition the output accumulator by channel rather than by query head, allowing all decoded values to be consumed without redundant decoding. Moreover, the affine transformation is folded into the attention contraction, so decoded levels can be used directly without explicit element-wise dequantization. Each rANS table entry jointly stores the reconstruction level and state-update information, allowing a single lookup to serve both entropy decoding and attention computation.
\section{Experiments}
\label{sec:experiments}

\subsection{Setup}
\label{sec:setup}

\paragraph{Models and benchmarks.} We evaluate D-Quant on Qwen3-8B \citep{yang2025qwen3} and Llama-3.1-8B-Instruct \citep{grattafiori2024llama3}. 
We evaluate on RULER \citep{hsieh2024ruler} at six context lengths from 4K to 128K and on the English subsets of LongBench-E \citep{bai2024longbench}, using greedy decoding for all evaluations.
RULER results are averaged over 100 samples per task at each context length.
Since Qwen3-8B is natively trained with a 40K context window, we apply static YaRN \citep{peng2024yarn} for evaluation at 64K and 128K. Although length extrapolation degrades the absolute performance at these lengths, all KV cache quantization methods are evaluated under the same setting, ensuring a fair comparison.

\paragraph{Baselines.} We compare D-Quant with KIVI \citep{liu2024kivi}, QuaRot \citep{ashkboos2024quarot}, TurboQuant \citep{zandieh2026turboquant}, and OScaR \citep{su2026oscar}, using BF16 FlashAttention-2 \citep{dao2023flashattention2} as the full-precision reference. We use the official implementations and recommended configurations whenever available. All quantized baselines use a group size of $128$. KIVI, QuaRot and OScaR use asymmetric INT2 quantization.
TurboQuant applies Lloyd--Max quantization to keys and affine quantization to values.
For TurboQuant, we use its MSE-only configuration, which is the applicable configuration at the evaluated storage budget.
KIVI, QuaRot, and TurboQuant retain the most recent $128$ tokens in BF16, while OScaR uses its recommended $256$-token residual window. Their effective storage costs are $2.25$, $2.25$, $2.19$, and $2.26$ bits per value, respectively, including quantization metadata such as scales and zero points.

\paragraph{Implementation details.} Unless otherwise specified, D-Quant allocates $331$-byte and $231$-byte containers to keys and values, with $8$ and $6$ quantization levels, respectively. This corresponds to an average storage cost of $2.26$ bits per value. Each token is quantized as a single group of $1024$ values, the entropy model uses a sharpening factor of $\alpha=1.4$, and the most recent $128$ tokens are retained in BF16. The same configuration is used across all models, layers, context lengths, and benchmarks. All experiments are conducted on NVIDIA H20 96GB GPUs.

\subsection{Main Results}
\label{sec:exp:main}

% Generated by DriftKV/experiments/make_main_table.py -- do not edit by hand.
% Source: accuracy.json, written by D-Quant/scripts/score_accuracy.py
\providecommand{\best}[1]{\textbf{#1}}
\definecolor{tblours}{HTML}{E8F1FA}

\begin{table*}[t]
\centering
\begingroup
\small
\caption{\textbf{Accuracy (\%) on LongBench-E and RULER.} \textbf{Bold} indicates the best method.}
\label{tab:main-longbench-ruler}
\setlength{\tabcolsep}{4pt}
\renewcommand{\arraystretch}{0.95}
\resizebox{\textwidth}{!}{%
\begin{tabular}{l c *{7}{c} !{\color{black!35}\vrule width 0.4pt} *{7}{c}}
\toprule
& & \multicolumn{7}{c}{\textbf{LongBench-E}}
& \multicolumn{7}{c}{\textbf{RULER}} \\
\cmidrule(lr){3-9}\cmidrule(lr){10-16}
\textbf{Method} & \textbf{Bit}
& \textbf{SQA} & \textbf{MQA} & \textbf{Sum} & \textbf{FS} & \textbf{Syn} & \textbf{Code} & \textbf{AVG}
& \textbf{4K} & \textbf{8K} & \textbf{16K} & \textbf{32K} & \textbf{64K} & \textbf{128K} & \textbf{AVG} \\
\midrule
\multicolumn{16}{c}{\rule{0pt}{2.3ex}\textbf{Qwen3-8B}} \\
\midrule
BF16 & 16.00 & 18.70 & 13.07 & 25.79 & 67.92 & 55.33 & 68.57 & 41.56 & 94.66 & 93.68 & 93.29 & 91.29 & 78.78 & 72.13 & 87.31 \\
\midrule
KIVI & 2.25 & 16.81 & 12.25 & 24.30 & 67.00 & 47.73 & 63.85 & 38.66 & 84.22 & 81.12 & 78.36 & 69.43 & 48.33 & 33.83 & 65.88 \\
QuaRot & 2.25 & 16.84 & 12.42 & 23.99 & 67.32 & 55.31 & 66.84 & 40.45 & 90.70 & 90.97 & 88.29 & 84.64 & 64.18 & 56.47 & 79.21 \\
TurboQuant & 2.19 & 14.57 & 12.02 & 20.43 & 61.91 & \best{55.95} & 53.97 & 36.48 & 76.32 & 69.63 & 63.49 & 65.28 & 31.27 & 21.18 & 54.53 \\
OScaR & 2.26 & 17.04 & 12.83 & 24.20 & 67.48 & 55.74 & 66.70 & 40.67 & 89.60 & 88.91 & 86.82 & 83.12 & 62.93 & 57.14 & 78.09 \\
\midrule
\rowcolor{tblours}\textbf{D-Quant}$_{\text{1.96}}$ & 1.96 & 18.05 & 13.18 & 25.30 & 67.21 & 52.32 & 67.75 & 40.64 & 93.90 & 92.37 & 92.85 & 90.11 & 74.38 & 67.67 & 85.21 \\
\rowcolor{tblours}\textbf{D-Quant} & 2.26 & \best{18.33} & \best{12.91} & \best{26.01} & \best{67.71} & 51.52 & \best{69.56} & \best{41.00} & \best{93.78} & \best{93.27} & \best{93.16} & \best{92.13} & \best{75.77} & \best{70.92} & \best{86.50} \\
\midrule
\multicolumn{16}{c}{\rule{0pt}{2.3ex}\textbf{Llama-3.1-8B}} \\
\midrule
BF16 & 16.00 & 19.96 & 17.68 & 30.66 & 67.29 & 57.38 & 62.03 & 42.50 & 95.69 & 95.82 & 94.70 & 89.51 & 85.96 & 76.44 & 89.69 \\
\midrule
KIVI & 2.25 & \best{19.92} & \best{24.19} & 28.95 & 66.31 & \best{57.50} & 56.88 & \best{42.29} & 87.24 & 83.24 & 79.68 & 75.37 & 69.18 & 51.87 & 74.43 \\
QuaRot & 2.25 & 18.26 & 15.29 & 25.49 & 64.41 & 46.07 & 47.45 & 36.16 & 80.68 & 76.01 & 71.24 & 65.53 & 60.74 & 38.15 & 65.39 \\
TurboQuant & 2.19 & 18.15 & 15.18 & 27.80 & 66.27 & 56.52 & 56.05 & 40.00 & 90.52 & 87.16 & 85.69 & 77.76 & 71.78 & 53.84 & 77.79 \\
OScaR & 2.26 & 19.18 & 17.12 & 30.00 & \best{67.81} & 57.48 & 59.58 & 41.87 & 93.98 & 93.27 & 90.97 & 85.40 & 80.30 & 66.15 & 85.01 \\
\midrule
\rowcolor{tblours}\textbf{D-Quant}$_{\text{1.96}}$ & 1.96 & 18.97 & 17.45 & 29.19 & 67.07 & 56.63 & 61.23 & 41.76 & 94.39 & 94.14 & 92.72 & 87.09 & 84.52 & 70.72 & 87.26 \\
\rowcolor{tblours}\textbf{D-Quant} & 2.26 & 19.01 & 17.27 & \best{30.41} & 67.73 & 56.56 & \best{61.59} & 42.10 & \best{95.59} & \best{94.78} & \best{93.97} & \best{89.10} & \best{85.53} & \best{74.65} & \best{88.94} \\
\bottomrule
\end{tabular}%
}
\endgroup
\end{table*}

\paragraph{D-Quant consistently preserves long-context performance.} As shown in Table~\ref{tab:main-longbench-ruler}, D-Quant at $2.26$ bits per value consistently preserves model quality across both models. On RULER, D-Quant achieves average scores of $86.50$ and $88.94$ on Qwen3-8B and Llama-3.1-8B, respectively, closely matching the BF16 references of $87.31$ and $89.69$ while substantially outperforming all low-bit baselines. On LongBench-E, where the differences among compressed methods are considerably smaller, D-Quant remains competitive and maintains performance comparable to BF16.

\paragraph{D-Quant remains robust as context length increases.} The performance gap between D-Quant and existing quantization methods becomes more pronounced at longer context lengths. While most baselines perform reasonably well at short contexts, their accuracy deteriorates rapidly as the sequence length increases. D-Quant, in contrast, consistently tracks the BF16 reference across the entire context range. At $128$K, D-Quant achieves $70.92$ and $74.65$ on Qwen3-8B and Llama-3.1-8B, respectively, while the strongest baselines reach only $57.14$ and $66.15$.

\paragraph{D-Quant remains strong at a lower storage budget.} D-Quant can further reduce the storage budget without being constrained by conventional fixed bit widths. At $1.96$ bits per value, it still achieves RULER averages of $85.21$ on Qwen3-8B and $87.26$ on Llama-3.1-8B, outperforming all baselines despite using less storage. This demonstrates that D-Quant enables more flexible trade-offs between storage and accuracy than fixed-width quantization methods.

\subsection{Efficiency}
\label{sec:exp:efficiency}

We evaluate the end-to-end efficiency of D-Quant on Qwen3-8B using a single NVIDIA H20 96GB GPU. All measurements use the actual compressed KV cache.
We consider two settings: (1) an $8$K prompt with increasing batch sizes for decoding throughput and peak GPU memory, and (2) batch size $1$ with a $1$K prompt and continuous decoding to $64$K tokens for long-context KV cache memory.

\begin{figure}[t]
\centering
\includegraphics[width=0.32\textwidth]{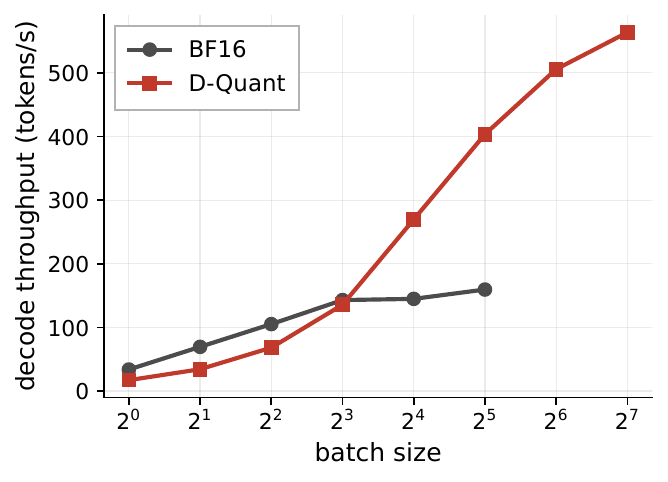}\hfill
\includegraphics[width=0.32\textwidth]{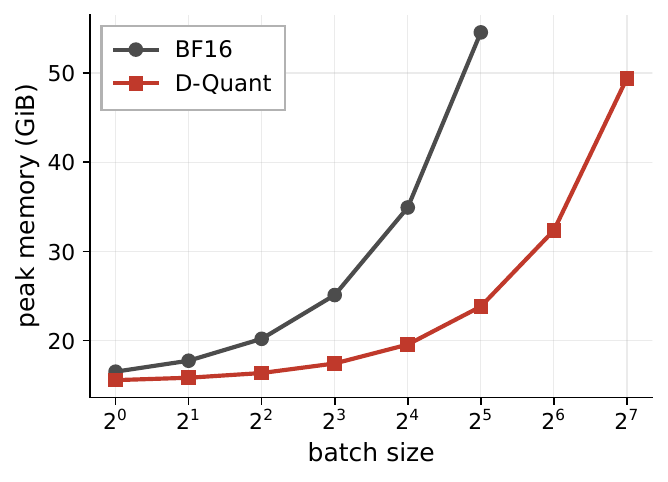}\hfill
\includegraphics[width=0.32\textwidth]{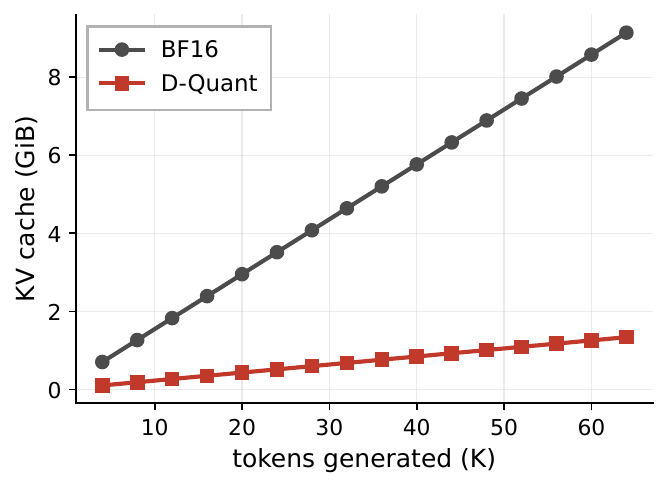}
\caption{\textbf{System efficiency on Qwen3-8B.} Left: decoding throughput with increasing batch size for an $8$K prompt. Middle: peak GPU memory under the same setting. Right: KV cache memory when decoding from a $1$K prompt to $64$K tokens at batch size $1$.}
\label{fig:efficiency}
\end{figure}

\paragraph{D-Quant substantially reduces KV cache memory.} As shown in Figure~\ref{fig:efficiency}, D-Quant significantly reduces memory consumption in both high-batch and long-context settings. With an $8$K prompt, BF16 runs out of memory at batch size $64$, whereas D-Quant scales to batch size $128$ with only $49.4$\,GiB peak memory. During long-context decoding, D-Quant reduces the KV cache memory by approximately $7\times$ at $64$K. The smaller memory footprint directly enables substantially higher serving capacity under a fixed GPU memory budget.

\paragraph{Higher capacity translates into higher serving throughput.} Entropy decoding introduces additional computation, making D-Quant slower than BF16 at small batch sizes. As the batch size increases, the reduced KV cache enables substantially higher throughput. At their respective maximum feasible batch sizes, D-Quant reaches $563.6$ tokens/s compared with $159.5$ tokens/s for BF16, yielding up to $3.5\times$ higher throughput. Overall, D-Quant trades additional decoding computation for substantially lower memory consumption and higher memory-constrained serving throughput.

\subsection{Analysis}
\label{sec:exp:ablations}

\begin{figure}[t]
\centering
\includegraphics[width=\textwidth]{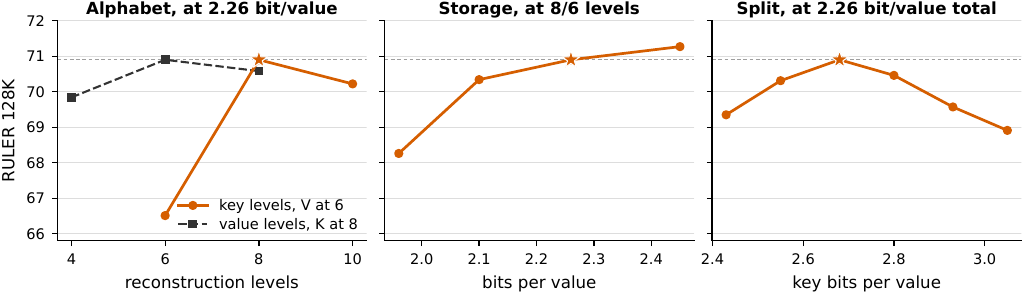}
\caption{\textbf{Ablation of D-Quant's main design choices on Qwen3-8B at 128K.} Each panel varies one configuration while keeping the others fixed. Left: quantization levels at a fixed $2.26$ bits per value. Middle: storage budget with $8/6$ key/value levels. Right: K/V budget allocation at a fixed total storage of $2.26$ bits per value. The star marks the default configuration.}
\label{fig:ablations}
\end{figure}

We explore the main design choices of D-Quant on RULER at $128$K with Qwen3-8B.

\paragraph{Larger alphabets improve representation until drift becomes costly.} In our main configuration, we allocate $2.68$ bits to keys and $1.84$ bits to values, which correspond to approximately $6$ and $3$ levels, respectively.
Since entropy coding compresses the non-uniform symbol distribution more efficiently, D-Quant can use larger alphabets under the same storage budget.
As shown in Figure~\ref{fig:ablations} (left), using $8$ levels for keys and $6$ for values provides a favorable operating point. Further increasing either alphabet makes drift more frequent to satisfy the container constraint, eventually degrading model performance.

\paragraph{Increasing the storage budget steadily approaches BF16 performance.} Figure~\ref{fig:ablations} (middle) shows that increasing the container size consistently improves accuracy, with the gain gradually saturating as performance approaches the BF16 reference. This provides D-Quant with a flexible trade-off between KV cache storage and model quality.

\paragraph{Keys benefit from a moderately larger budget.} Most KV cache quantization methods allocate the same storage budget to keys and values. As shown in Figure~\ref{fig:ablations} (right), assigning moderately more capacity to keys improves performance, consistent with their higher sensitivity to quantization error. However, overly skewing the budget toward keys eventually hurts accuracy as the value representation becomes too constrained. We therefore allocate slightly more bits to keys in our method.
\section{Conclusion}

We present D-Quant, an entropy-coded KV cache quantization framework that combines the compression efficiency of entropy coding with the regular memory layout required for efficient inference.
D-Quant encodes each KV stream into a fixed-size container and uses length-constrained drift to ensure that every stream fits its storage budget.
Experiments across models and long-context benchmarks show that D-Quant consistently preserves near-BF16 performance while substantially reducing KV cache memory and improving serving throughput.

\bibliography{iclr2027_conference}
\bibliographystyle{iclr2027_conference}

\newpage
\appendix

\section{Encoding and Decoding}
\label{app:algorithm}

This section provides the detailed encoding and decoding procedures omitted from the main text. Algorithm~\ref{alg:drift} gives the encoder for one stream. Two implementation details are worth noting. First, we bisect the exponent of the Lagrange multiplier rather than the multiplier itself, which provides stable convergence across the dynamic range of distortion values. Second, the scale and offset are rounded to FP16 before the final assignment, so that the encoder optimizes against exactly the reconstruction grid used by the decoder.

\begin{algorithm}[h]
\caption{Encoding one stream}
\label{alg:drift}
\begin{algorithmic}[1]
\Require values $\vx \in \mathbb{R}^{n}$, frequency table $\vf$ of size $M$, container $C$ bytes
\State $B \gets 8C-\sigma$;\quad
       $r_m \gets -\log_2(f_m/\textstyle\sum_j f_j)$;\quad
       $s,o \gets \textsc{FitMinMax}(\vx,M)$
\State $\vm \gets \textsc{Drift}(\vx,s,o,\vr,B)$
\State $s,o \gets \textsc{Refit}(\vx,\vm)$
\State $s,o \gets \textsc{fp16}(s),\textsc{fp16}(o)$
\State $\vm \gets \textsc{Drift}(\vx,s,o,\vr,B)$
\State \Return $\textsc{rANSEncode}(\vm,\vf,C),\,s,\,o$
\Statex
\Function{Drift}{$\vx,s,o,\vr,B$}
  \State $D_{im} \gets (x_i-(o+s\,m))^2$
  \State $\text{hi} \gets \log_2\!\bigl(8(\max D-\min D)\bigr)$;\quad
         $\text{lo} \gets \text{hi}-40$
  \For{$1$ to $T$} \Comment{$T=12$; bisect $\log_2\lambda$}
    \State $\mu \gets (\text{lo}+\text{hi})/2$
    \State $\vm \gets \arg\min_m\left(D_{im}+2^\mu r_m\right)$
    \State \textbf{if} $\sum_i r_{m_i}\le B$ \textbf{then}
           $\text{hi}\gets\mu$
           \textbf{else} $\text{lo}\gets\mu$
  \EndFor
  \State \Return $\arg\min_m\left(D_{im}+2^{\text{hi}}r_m\right)$
\EndFunction
\end{algorithmic}
\end{algorithm}

Algorithm~\ref{alg:decode} gives the corresponding rANS decoder. The stream is encoded in reverse order so that decoding produces symbols in their original order. With $p=8$, the low eight bits of the state directly index a $256$-entry table. Each entry stores the reconstruction level together with the frequency and cumulative offset required for the state update, allowing entropy decoding and reconstruction to share a single lookup.

\begin{algorithm}[h]
\caption{Decoding one stream}
\label{alg:decode}
\begin{algorithmic}[1]
\Require payload of $C$ bytes, table $T[z]=(v_m,f_m,c_m)$
\State $x \gets \textsc{LittleEndian}(\text{payload}[0{:}\sigma/8])$;\quad
       $\text{ptr}\gets\sigma/8$
\For{$i=1$ \textbf{to} $n$}
  \State $z \gets x \bmod 2^p$
  \State $(v_{m_i},f_{m_i},c_{m_i}) \gets T[z]$
  \State $x \gets f_{m_i}\lfloor x/2^p\rfloor+z-c_{m_i}$
  \While{$x<L$}
    \State $x \gets (x\ll8)\mathbin{|}\text{payload}[\text{ptr}]$;\quad
           $\text{ptr}\gets\text{ptr}+1$
  \EndWhile
  \State \textsc{Consume}$(i,v_{m_i})$
\EndFor
\end{algorithmic}
\end{algorithm}

The decoded levels are consumed directly by the attention kernel rather than materialized as reconstructed KV tensors. For a key stream of token $t$, the reconstructed channel $j$ is
\begin{equation}
    \hat{k}_{t,j}=s_t v_{m_{t,j}}+o_t+\mu_j ,
\end{equation}
where $\vmu$ is the windowed channel-wise mean. Since $s_t$ and $o_t$ are constant across the channel contraction,
\begin{equation}
    \vq^\top\hat{\vk}_t
    =
    s_t\sum_j q_jv_{m_{t,j}}
    +
    o_t\sum_j q_j
    +
    \sum_j q_j\mu_j .
    \label{eq:fold-key}
\end{equation}
The inner product can therefore accumulate decoded reconstruction levels directly and apply the affine terms after the contraction.

For values, the contraction is instead performed across tokens. For output channel $j$,
\begin{equation}
    \sum_t a_t\hat{v}_{t,j}
    =
    \sum_t (a_t s_t)v_{m_{t,j}}
    +
    \sum_t a_t o_t ,
    \label{eq:fold-value}
\end{equation}
where $a_t$ is the attention weight for token $t$. The per-token scale is folded into the corresponding attention weight, while the offset is accumulated separately. Thus, neither path requires explicit element-wise dequantization or materialization of a BF16 KV cache. Since every stream occupies exactly $C$ bytes, its address is determined directly from the token index without an offset table or dynamic allocator.

\section{Comparison with Fixed-Width Coding at Matched Storage}
\label{app:matched}

This experiment isolates D-Quant's coding mechanism from its preprocessing. We keep the preprocessing, grouping, and storage budget unchanged and compare the entropy-coded container with a fixed-width representation at matched storage.

\paragraph{Setting.} We use $27$ held-out LongBench-E documents per model, with three documents sampled from each of nine subsets. Each document provides a $2{,}048$-token context. We replace the cached keys and values of every layer with their reconstructed counterparts and measure the token-wise forward KL divergence from the unquantized model over the following $256$ positions. We report the geometric mean of the per-document KL ratios, with $95\%$ confidence intervals obtained from $20{,}000$ paired bootstrap resamples.

\paragraph{Compared configurations.} Both configurations use identical Hadamard rotation, windowed INT8 channel-mean removal for keys, token-wise min--max affine grids, per-token grouping, least-squares refitting, and FP16 metadata. D-Quant uses the default $331$/$231$-byte key/value containers with $8$/$6$ levels, rANS, and drift. The fixed-width counterpart receives the same storage budget and is conservatively charged the ideal $\log_2 M$ bits per symbol, which permits $6$ key levels and $3$ value levels; symbols are assigned by nearest-neighbor rounding.

\begin{table}[h]
\centering
\small
\caption{\textbf{D-Quant versus fixed-width coding at matched storage.} Both use identical preprocessing and grouping. KL is measured against the unquantized model}
\label{tab:matched}
\begin{tabular}{llcccc}
\toprule
Model & Quantity & D-Quant & Fixed width & Ratio ($95\%$ CI) & Worse \\
\midrule
\multirow{3}{*}{Qwen3-8B}
& KL divergence & $0.0312$ & $0.0705$ & $2.26\times$ $[2.09,2.46]$ & $27/27$ \\
& NMSE, key     & $1.22\%$ & $2.07\%$ & $1.70\times$ & $27/27$ \\
& NMSE, value   & $14.31\%$ & $39.98\%$ & $2.79\times$ & $27/27$ \\
\midrule
\multirow{3}{*}{Llama-3.1-8B}
& KL divergence & $0.0410$ & $0.0818$ & $2.00\times$ $[1.85,2.16]$ & $27/27$ \\
& NMSE, key     & $2.54\%$ & $4.34\%$ & $1.71\times$ & $27/27$ \\
& NMSE, value   & $14.82\%$ & $41.49\%$ & $2.80\times$ & $27/27$ \\
\bottomrule
\end{tabular}
\end{table}

\begin{figure}[h]
\centering
\includegraphics[width=\textwidth]{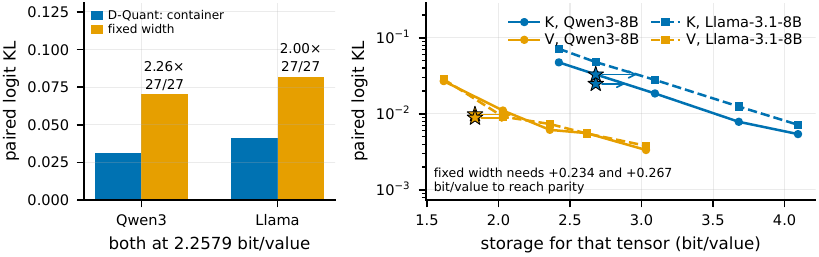}
\caption{\textbf{D-Quant versus fixed-width coding at matched storage.} Left: paired divergence at the default storage budget. Right: divergence as the storage budget for each tensor increases. Stars mark the default D-Quant operating points, and arrows indicate the additional storage required by fixed-width coding to reach comparable divergence.}
\label{fig:matched_rate}
\end{figure}

\paragraph{D-Quant achieves lower distortion at the same storage.} As shown in Figure~\ref{fig:matched_rate}, D-Quant approximately halves the divergence from the full-precision model at matched storage, with fixed-width coding performing worse on every evaluated document. Conversely, matching D-Quant's divergence requires fixed-width coding to spend an additional $0.234$ and $0.267$ bits per value on Qwen3-8B and Llama-3.1-8B, respectively. This confirms that the improvement is not due to preprocessing or additional storage, but to using the available storage more effectively.

\section{Empirical Analysis of the Coding Mechanism}
\label{app:mechanism}

\begin{figure}[h]
\centering
\includegraphics[width=\textwidth]{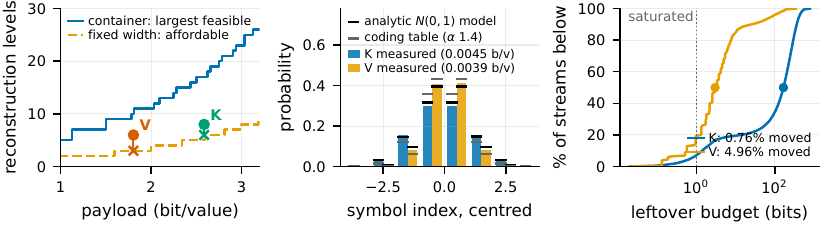}
\caption{\textbf{Empirical analysis of D-Quant's coding mechanism on Qwen3-8B.} Left: maximum feasible alphabet size under the entropy-coded container and equal-rate fixed-width coding. Middle: analytic probability model, sharpened coding table, and measured symbol distributions. Right: distribution of the remaining container budget after drift; only streams near the storage limit require appreciable reassignment. Results are collected from six layers spanning the model depth.}
\label{fig:mechanism}
\end{figure}

\paragraph{Entropy coding supports a substantially richer alphabet.} Figure~\ref{fig:mechanism} (left) shows the maximum number of reconstruction levels supported as the payload budget varies. Fixed-width coding increases the alphabet only when the available rate crosses the next code-width threshold, whereas the entropy-coded container can exploit the concentrated symbol distribution to support substantially more levels at the same rate. The default $8/6$ configuration lies comfortably within this feasible region, providing additional reconstruction resolution while leaving sufficient capacity for drift.

\paragraph{The analytic entropy model closely matches the measured distribution.} Figure~\ref{fig:mechanism} (middle) compares the calibration-free probability model with the measured symbol distributions. Despite using no calibration data, the analytic model incurs only $0.0045$ and $0.0039$ additional bits per value for keys and values, respectively. The final coding table is intentionally sharper than this model: drift tends to move symbols toward more probable central levels, and sharpening anticipates the resulting coded distribution.

\paragraph{Drift makes only sparse, local corrections.} Figure~\ref{fig:mechanism} (right) shows that many streams already have substantial remaining budget, while drift becomes relevant for streams approaching the container limit. Across all KV streams, only $0.75\%$ of key symbols and $5.00\%$ of value symbols are changed, and no changed symbol moves by more than one reconstruction level. Thus, drift satisfies the hard per-stream constraint through sparse local adjustments rather than globally reducing the quantization resolution.
\end{document}